\documentclass[conference]{IEEEtran}
\IEEEoverridecommandlockouts
\usepackage{cite}
\usepackage{amsmath,amssymb,amsfonts}
\usepackage{algorithmic}
\usepackage{graphicx}
\usepackage{textcomp}
\usepackage{xcolor}
\usepackage{hyperref} 
\usepackage{booktabs}
\usepackage{multirow}
\usepackage{etoolbox}
\usepackage{utfsym}
\hypersetup{hidelinks}

\def\BibTeX{{\rm B\kern-.05em{\sc i\kern-.025em b}\kern-.08em
    T\kern-.1667em\lower.7ex\hbox{E}\kern-.125emX}}

\begin{document}

\title{SGF-CDNet: A Consistency-Discrepancy Graph Network over Semantic-Geometric Fused Nodes for Face Forgery Detection\vspace{-0.3em}}


\author{
\parbox{16cm}{\centering
    {\large Jiayao Jiang\textsuperscript{1,2}, Bin Liu\textsuperscript{1,2\dag}, Nenghai Yu\textsuperscript{1,2}}\\
    {\normalsize
    \textsuperscript{1}School of Cyber Science and Technology, University of Science and Technology of China, China\\
    \textsuperscript{2}Anhui Province Key Laboratory of Digital Security, China\\[-1.5em]
    \thanks{\dag Corresponding author: \href{mailto:flowice@ustc.edu.cn}{flowice@ustc.edu.cn}}}
}}

\maketitle


\begin{abstract}
The rapid advancement of deepfakes necessitates robust face forgery detection. Although forged faces may lack obvious artifacts, they often contain subtle disharmony among different facial regions. We propose SGF-CDNet, a Consistency-Discrepancy Graph Network (CD-GNN) over Semantic-Geometric Fused (SGF) nodes. First, SGF-CDNet constructs SGF nodes by deeply fusing semantic regions from face parsing with geometric information from facial landmarks, allowing nodes to capture both high-level concepts and precise geometric constraints. Next, a dual-path CD-GNN performs parallel relational reasoning on these nodes across two dimensions: consistency and discrepancy. The consistency path evaluates if facial components follow natural biological patterns, while the discrepancy path mines for structural tensions and feature conflicts introduced by forgeries. By integrating these processes, our model effectively identifies disharmonious relationships between facial components. Extensive experiments on public datasets demonstrate that SGF-CDNet achieves superior performance, establishing it as a reliable solution for face forgery detection.
\end{abstract}


\section{Introduction}
\label{sec:intro}


The rapid development of deepfake generation technologies poses serious threats to social trust and personal information security. Therefore, robust and reliable face forgery detection algorithms are urgently needed. Currently, mainstream deep learning detectors have achieved great success in identifying specific forgery types, but their generalization ability tends to decrease when facing a continuous stream of increasingly advanced forgery techniques. This highlights the limitations of existing methods in coping with unknown attacks.

\begin{figure}[thpb]
  \centering
\includegraphics[width=7.5cm]{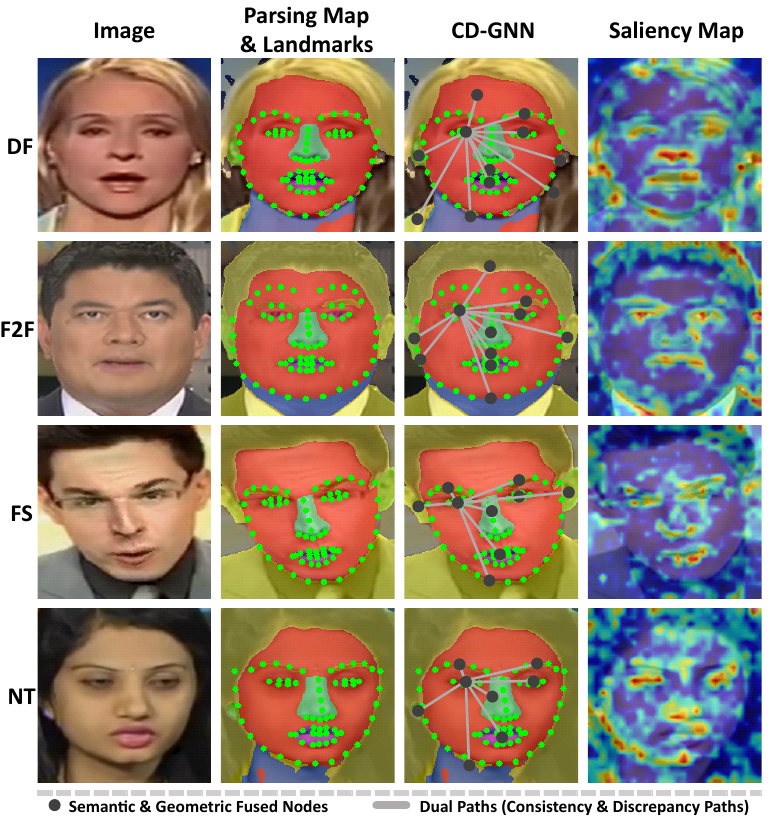}
  \vspace{-6mm}
  \caption{Example frames, face parsing results and landmarks, subgraphs of CD-GNN and saliency maps with four different manipulation types: DeepFakes (DF), Face2Face (F2F), FaceSwap (FS) and NeuralTextures (NT) from FaceForensics++ (FF++) dataset \cite{Rossler_2019_ICCV}.}
  \label{figure1}
\vspace{-7mm}
\end{figure}

As shown in the first column of Fig. \ref{figure1}, some manipulated images do not exhibit obvious artifacts, yet they may have the unnatural and disharmonious relationship between different regions of faces. As a result, some studies have begun to explore the use of Graph Neural Networks (GNNs) \cite{4700287} to model the structural relationships between facial components. However, when constructing graph nodes, these methods either rely on sparse landmarks \cite{yan2023multimodalgraphlearningdeepfake}, thus lacking regional semantics, or utilize semantic regions but fail to deeply fuse them with precise geometric information \cite{articleZou}, leading to an incomplete information representation for the graph nodes. Moreover, their graph reasoning processes are often simplistic, failing to explore the potential structural-level disruptions to harmony and the internal conflicts introduced by the forgeries.

\begin{figure*}[thpb]
  \centering 
  \includegraphics[width=0.85\textwidth]{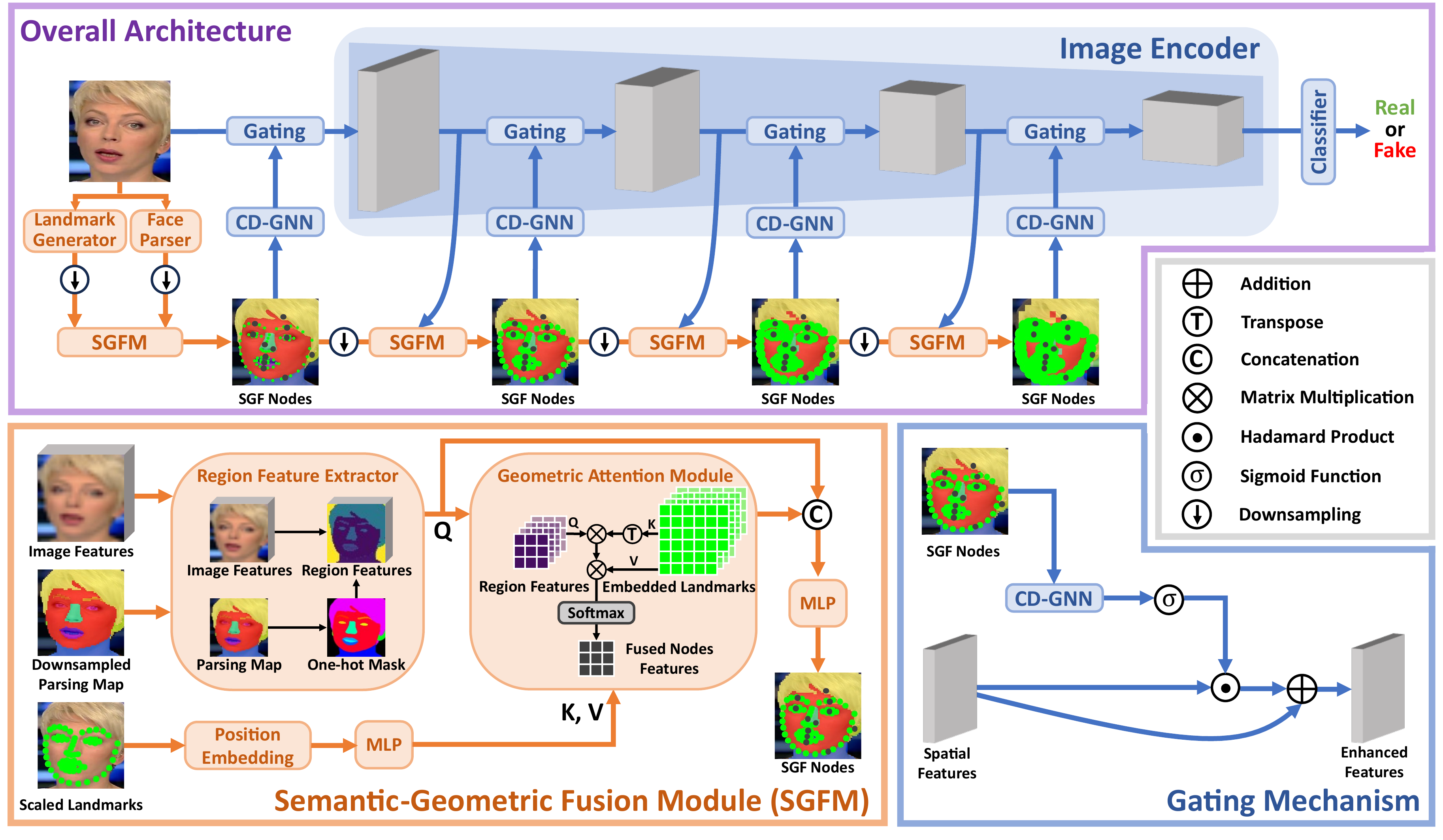}
  \vspace{-4mm}
  \caption{Overview of model architechture with SGF-CDNet. The proposed SGF-CDNet employs a Semantic-Geometric Fusion Module (SGFM) to build SGF nodes from facial semantic and geometric information and a Consistency-Discrepancy Graph Network (CD-GNN, detailed in Fig. \ref{figure3}) to perform relational reasoning. The output of SGF-CDNet is applied to downsampled features from four different stages of the image encoder through gating mechanisms.}
  \label{figure2}
\vspace{-6mm}
\end{figure*}

To address the limitations of current GNN-based models, we propose an innovative Consistency-Discrepancy Graph Network over Semantic-Geometric Fused nodes for face forgery detection (SGF-CDNet). Our framework first constructs Semantic-Geometric Fused (SGF) nodes that possess both high-level semantics and precise geometry by fusing the information of face parsing results and landmarks, as shown in column 2 of Fig. \ref{figure1}. Subsequently, a novel dual-path Consistency-Discrepancy Graph Network (CD-GNN) performs relational reasoning in parallel on the relationships between these SGF nodes to effectively capture natural and unnatural structural association with facial components, as shown in column 3 of Fig. \ref{figure1}. This dual path design is a key part of our approach. The consistency path checks whether facial components follow natural and biologically correct patterns. In contrast, the discrepancy path actively finds and highlights the unnatural structural tensions and feature conflicts that are typical signs of manipulations. By analyzing the face from these two different but complementary angles at once, our model can detect manipulations more effectively. The main contributions of this paper are as follows:
\begin{itemize}
\item To address the limitations of current GNN-based methods, We propose a novel dual-path Consistency-Discrepancy Graph Network (CD-GNN) over Semantic-Geometric Fused (SGF) nodes for face forgery detection (SGF-CDNet) based on structured graph reasoning. 
\item We design an SGF nodes construction method that fuses facial semantic regions with geometric landmarks, significantly enhancing the information density and representational capacity of the graph nodes.
\item We propose a novel CD-GNN framework that can perform deeper and more comprehensive relational reasoning on the facial structure from complementary dimensions, including consistency and discrepancy. These dimensions can better capture the harmonious and disharmonious relationships between facial components. 
\item Experimental results on public datasets show that our method achieves excellent performance. Additionally, extensive ablation studies further improve the effectiveness of our design components.
\end{itemize}

\section{METHODOLOGY}
\vspace{-1mm}
\subsection{Architecture Overview}
\vspace{-1mm}
The overall architecture of SGF-CDNet is shown in Fig. \ref{figure2}. This framework consists of three components: a multi-scale image encoder, a Semantic-Geometric Fusion Module (SGFM) and a Consistency-Discrepancy Graph Network (CD-GNN). Specifically, the input image first passes through the image encoder to extract hierarchical feature maps at different stages. At each stage, the SGFM module constructs nodes with rich context information by leveraging features at certain scale, combined with semantic information from face parsing results and geometric information from facial landmarks. Then, these nodes are fed into the CD-GNN for structured relational reasoning. The CD-GNN analyzes the relationships between nodes via two parallel paths, from the complementary perspectives of consistency and discrepancy, to uncover the structural disorder that is introduced by disharmonious relationships between facial components. Finally, the structured information inferred by graph network enhances the encoder's original feature maps through a gating mechanism adaptively. This process is worked at multiple scales.
\vspace{-2mm}
\subsection{Semantic-Geometric Fusion Module}
\vspace{-1mm}
The expressive power of GNN largely depends on the quality of its nodes. To construct SGF nodes that can simultaneously represent high-level semantic concepts and precise geometric constraints, we designed the Semantic-Geometric Fusion Module (SGFM). This module aims to deeply fuse discrete semantic regions with landmark coordinates, providing a dense information input for the subsequent graph reasoning. The detailed workflow of the SGFM is illustrated in Fig. \ref{figure2}, which consists of two core components: region feature extractor and geometric attention module.
\subsubsection{Region Feature Extractor}
This stage aims to extract feature vectors from the image feature map corresponding to specific facial semantic regions, such as the eyes and nose. For the feature map $F_i \in \mathbb{R}^{C_i \times H_i \times W_i}$ output by the image encoder at stage $i$ and the corresponding downsampled face parsing map $M_i \in \{0, 1, ..., N\}^{H_i \times W_i}$, we first convert $M_i$ into a set of one-hot encoded masks $\{M_{i,n}^{\text{onehot}} \in \{0, 1\}^{H_i \times W_i}\}_{n=1}^{N}$, where $N$ is the number of facial semantic regions. $C_i$, $H_i$ and $W_i$ are the number of channels, height and width of the feature map $F_i$, respectively. Subsequently, the initial feature $v_n^\text{sem}$ for each semantic region $n$ is obtained by applying weighted average pooling to the feature map $F_i$:
\begin{equation}
     v_n^\text{sem} = \frac{\sum_{h,w} F_i(h,w) \cdot M_{i,n}^\text{onehot}(h,w)}{\sum_{h,w} M_{i,n}^\text{onehot}(h,w) + \epsilon}, 
\label{1}
\end{equation}
where $h$ and $w$ represent the indices of row and column on the feature map, respectively. $\epsilon$ is a small value to prevent the denominator from being zero. In (\ref{1}), the average of all features within a specific facial region is calculated using a mask to obtain a single feature vector that represents the region. Thus, we obtain a set of semantic features purely based on regional content $V_\text{sem} = \{v_1^\text{sem}, \dots, v_N^\text{sem}\}$.

\begin{figure*}[thpb]
  \centering 
  \includegraphics[width=0.85\textwidth]{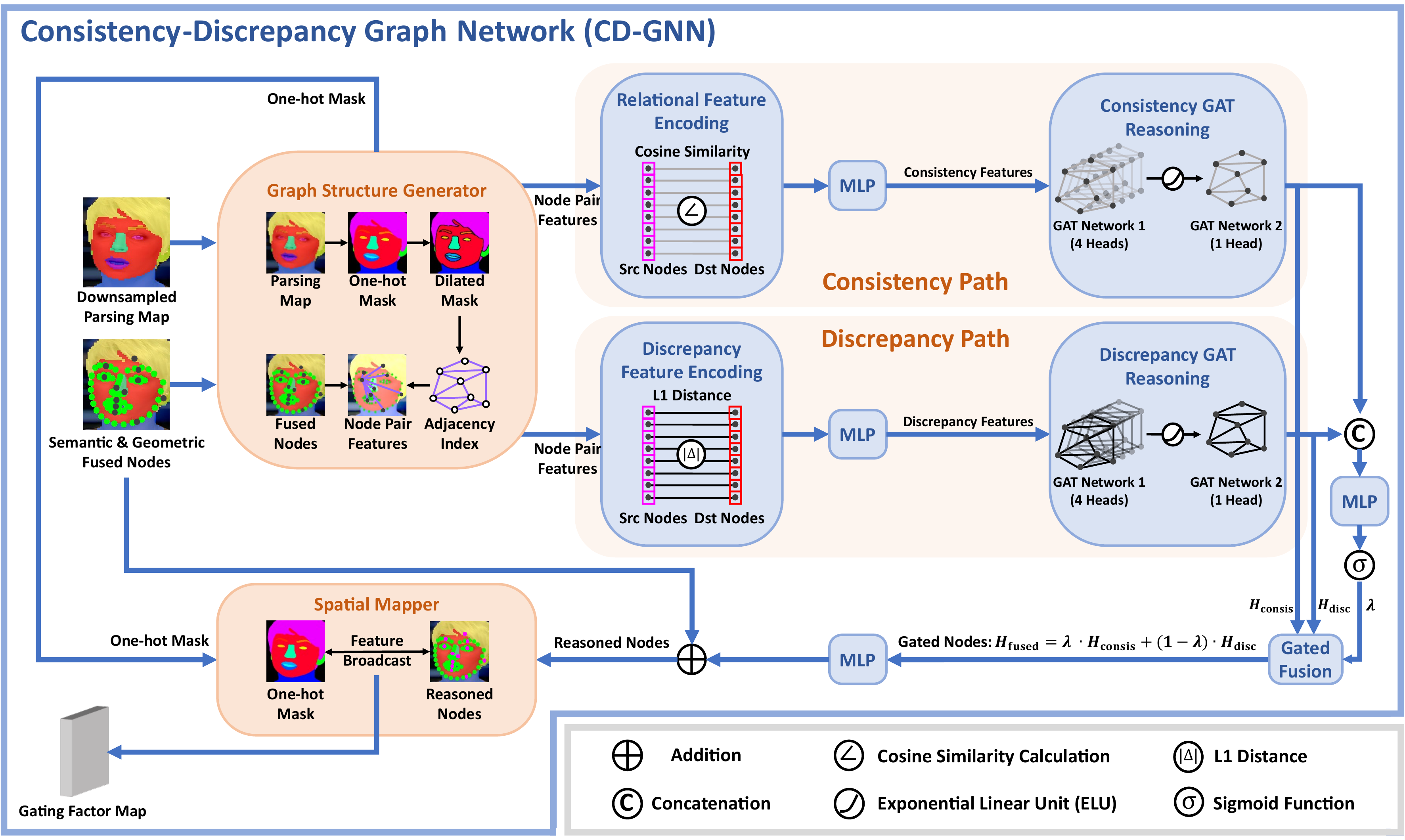}
  \vspace{-4mm}
  \caption{Detailed architecture of the CD-GNN. First, the graph structure generator determines the adjacency relationship between nodes. Subsequently, the fused node features are fed into two parallel reasoning paths. The outputs from the two paths are merged via gated fusion mechanism and combined with a residual connection to generate the reasoned nodes. Finally, the spatial mapper broadcasts these reasoned results back to the 2D space, creating the final gating factor map to guide feature enhancement.}
  \label{figure3}
  \vspace{-6mm}
\end{figure*}

\subsubsection{Geometric Attention Module}
To enable semantic features to capture their position within the overall facial geometric structure, we introduce a geometric attention module. First, $K$ correspondingly scaled facial landmark coordinates $L_i \in \mathbb{R}^{K \times 2}$ are encoded through sinusoidal positional embedding (PosEmbed) and a multi-layer perceptron (MLP) to obtain high dimensional geometric features $E_\text{geo} \in \mathbb{R}^{K \times C_{i}}$:
\begin{equation}
  E_\text{geo} = \text{MLP}(\text{PosEmbed}(L_i)).
\end{equation}

Next, we employ a cross-attention mechanism to fuse these two types of information. Specifically, the regional semantic features $V_\text{sem}$ serve as the queries, while the geometric features of all landmarks $E_\text{geo}$, act as both the keys and values. Through this mechanism, each semantic region adaptively aggregates the most relevant parts from the global geometric information, generating geometry-enhanced features $V_\text{fused}$: 
\begin{equation}
    V_\text{fused} = \text{softmax}\left(\frac{V_\text{sem}W_\text{sem}^Q(E_\text{geo}W_\text{geo}^K)^T}{\sqrt{d_k}}\right) (E_\text{geo}W_\text{geo}^V),
\end{equation}
where $d_k$ is the dimension of the key vector to scale the results of the dot product for training stability. $W_\text{sem}^Q$, $W_\text{geo}^K$ and $W_\text{geo}^V$ are the projection matrices for $V_\text{sem}$ and $E_\text{geo}$.

\subsubsection{Final Node Generation}
Finally, to combine original semantic information with geometry-enhanced information, we concatenate initial regional features $V_\text{sem}$ with fused features $V_\text{fused}$ from the attention module along feature dimension and then fuse them through an MLP to generate final node features $Z = \{z_1, \dots, z_N\}$ for graph construction:
\begin{equation}
     z_n = \text{MLP}(\text{Concat}(v_n^\text{sem}, v_n^\text{fused})), 
\end{equation}
where $n$ is the $n$-th node feature. These nodes $Z$ contain both regional semantic and texture information and incorporate precise global geometric constraints, establishing a solid foundation for subsequent graph reasoning.

\vspace{-2mm}
\subsection{Consistency-Discrepancy Graph Network}
\vspace{-1mm}
After obtaining high-quality node features $Z = \{z_1, \dots,$ $ z_N\}$, we designed the dual-path Consistency-Discrepancy Graph Network (CD-GNN) for deep relational reasoning, with its detailed architecture illustrated in Fig. \ref{figure3}. The core idea of this network is to analyze the relationships between nodes from complementary perspectives of consistency and discrepancy through two parallel paths.

\subsubsection{Graph Structure Generation}
We first construct the graph based on the spatial adjacency of the facial semantic regions. For the set of one-hot encoded masks $\{M_{i,n}^{\text{onehot}}\}_{n=1}^{N}$, which correspond to the downsampled face parsing map $M_i$, we apply a dilation operation using a $3\times3$ convolution kernel to obtain the dilated masks $\{M_{i,n}^{\text{dilated}}\}_{n=1}^{N}$. Subsequently, the adjacency matrix $A \in \{0,1\}^{N \times N}$ is generated by calculating the overlap between dilated masks and original masks:
\begin{equation}
    A_{nm} = \begin{cases} 1 & \text{if } \sum\limits_{h,w} M_{i,n}^\text{dilated}(h,w) \cdot M_{i,m}^\text{onehot}(h,w) > 0 
    \\ 0 & \text{otherwise} \end{cases}, 
\end{equation}
where $n$ and $m$ are the node indices in the adjacency matrix $A_{nm}$, representing the $n$-th and $m$-th semantic region nodes in the graph respectively. The adjacency matrix $A$ defines the edge set $\mathcal{E}$, thereby constructing an undirected graph $\mathcal{G} = (\mathcal{V}, \mathcal{E})$, where the node set is $\mathcal{V}=Z$. The edge set $\mathcal{E}$ specifies the node pair features for relational reasoning, meaning each edge $(z_n,z_m) \in \mathcal{E}$ corresponds to a set of inputs composed of source and target node features, which is then fed into the subsequent reasoning paths.

\subsubsection{Consistency Path} 
For any pair of nodes $(z_n,z_m) \in \mathcal{E}$ connected by an edge in the graph $\mathcal{G}$, we feed their features into two parallel reasoning paths: consistency path and discrepancy path.
The consistency path aims to model the natural, harmonious biological association patterns among facial components. First, in the relational feature encoding stage, we calculate the cosine similarity $s_{nm}$ for each pair of adjacent nodes $(z_n,z_m)$:
\begin{equation}
    s_{nm} = \frac{z_n \cdot z_m}{|z_n| |z_m|}.
\end{equation}

These similarity scores aim to quantify harmony of edges. They are then aggregated back to each source node $z_n$ to form a scalar $\bar{s}_n$, representing the node's overall harmony with its neighborhood. This scalar is then encoded via an MLP into a high-dimensional consistency feature $z_n^\text{consis}$:
\begin{equation}
\begin{aligned}
      \bar{s}_n &= \frac{1}{|\mathcal{N}(n)|} \sum\limits_{m \in \mathcal{N}(n)} s_{nm},\\
     \quad z_n^\text{consis} &= \text{MLP}(\bar{s}_n),
\end{aligned}
\end{equation}
where $\mathcal{N}(n)$ denotes the set of neighboring nodes for node $n$. These consistency features $Z_{\text{consis}} = \{z_1^{\text{consis}}, \dots, z_N^{\text{consis}}\}$ are then fed into a two-layer consistency graph attention network (GAT). The first GAT layer employs $H$ attention heads and the output of the $h$-th head is $Z'(h)$. These outputs are concatenated after an exponential linear unit (ELU) activation function to form intermediate features $Z'$. The second GAT layer then uses a single head to aggregate information from $Z'$. The final output is consistency representation $H_\text{consis}$. The whole process can be expressed as:
\begin{equation}
\begin{aligned}
Z'{(h)} &= \text{ELU}(\text{GAT}_{\text{consis}, 1}^{(h)}(Z_{\text{consis}})), \\
Z' &= \text{Concat}(Z'{(1)}, \dots, Z'{(H)}), \\
H_{\text{consis}} &= \text{GAT}_{\text{consis}, 2}(Z').
\end{aligned}
\end{equation}

\subsubsection{Discrepancy Path} The goal of this path is to actively find and amplify the unnatural conflicts introduced by forgeries. In the discrepancy feature encoding stage, we calculate the L1 distance for a node pair $(z_n,z_m)$ and process it through an MLP to obtain a discrepancy feature $d_{nm}$:
\begin{equation}
    d_{nm} = \text{MLP}(|z_n - z_m|).
\end{equation}

This discrepancy feature $d_{nm}$ represents the conflict signal and is aggregated back to the source node $z_n$ to form a conflict-oriented node representation $z_n^\text{disc}$. Subsequently, these representations are fed into a discrepancy graph attention network, which has independent parameters, to learn a global representation  $H_\text{disc}$ focused on facial structural anomalies through a similar two-layer GAT architecture with consistency graph attention network.

\begin{table*}[t]	
	\centering\renewcommand\arraystretch{1.2}\setlength{\tabcolsep}{12.5pt}
	\belowrulesep=0pt\aboverulesep=0pt
	\caption{ 
    Cross-dataset evaluation of various face forgery detection methods on the CDF2, DFDC, DFDCP and FFIW datasets. Results for baseline methods are directly quoted from their original publications. The best performance is highlighted in \textbf{BOLD} and the second best is \underline{underlined}.
    \vspace{-3mm}
    \label{cross-data}}
	\begin{tabular}{c|c|c|cc|cccc}
		\toprule
		\multirow{2}{*}{Method} & 
		\multirow{2}{*}{Venue} & 
		\multirow{2}{*}{Input Type} &
		\multicolumn{2}{c|}{Training Set} & \multicolumn{4}{c}{Test Set AUC (\%)}\\
		\cmidrule(lr){4-5}\cmidrule(lr){6-9}
		&\multicolumn{1}{c|}{} 
		&\multicolumn{1}{c|}{} 
		&\multicolumn{1}{c}{Real} 
		&\multicolumn{1}{c|}{Fake} 
		&\multicolumn{1}{c}{CDF2} 
		&\multicolumn{1}{c}{DFDC} 
		&\multicolumn{1}{c}{DFDCP}
		&\multicolumn{1}{c}{FFIW} \\
		\midrule
		SBI \cite{Shiohara_2022_CVPR} & CVPR 2022 & Frame & \usym{2713} &  & 93.18 & 72.42 & 86.15 & {84.83} \\
		SeeABLE \cite{Larue_2023_ICCV} & ICCV 2023 & Frame & \usym{2713} &  & 87.30 & 75.90 & 86.30 & - \\
		AUNet \cite{Bai_2023_CVPR} & CVPR 2023 & Frame & \usym{2713} &  & 92.77 & 73.82 & 86.16 & 81.45 \\
		LAA-Net \cite{Nguyen_2024_CVPR} & CVPR 2024 & Frame & \usym{2713} &  & 95.40 & - & 86.94 & - \\
		RAE \cite{10.1007/978-3-031-72943-0_23} & ECCV 2024 & Frame & \usym{2713} &  & 95.50 & {80.20} & 89.50 & - \\
		FreqBlender \cite{zhou2024freqblender} & NeurIPS 2024 & Frame & \usym{2713} &  & 94.59 & 74.59 & 87.56 & 86.14 \\
		UDD \cite{fu2025exploring} & AAAI 2025 & Frame & \usym{2713} & \usym{2713} & 93.10 & 81.20 & 88.10 & - \\
        WMamba \cite{peng2025wmamba} & ACM MM 2025 & Frame & 
        \usym{2713} &  & \underline{96.29} & \underline{82.97} & \underline{89.62} & \underline{86.59}\\
		\midrule
		TALL++ \cite{xu2024towards} & IJCV 2024 & Video & \usym{2713} & \usym{2713} & 91.96 & - & 78.51 & - \\
		NACO \cite{zhang2025learning} & ECCV 2024 & Video & \usym{2713} & \usym{2713} & 89.50 & - & 76.70 & - \\
        DFD-FCG \cite{11093395} & CVPR 2025 & Video & \usym{2713} & \usym{2713} & 95.00 & 81.81 & - & - \\
		\midrule
		SGF-CDNet (Ours) & ICMEW 2026 & Frame & \usym{2713} &  & \textbf{97.37} & \textbf{84.72} & \textbf{89.77} & \textbf{87.23} \\      
		\bottomrule
	\end{tabular}
    \vspace{-7mm}
\end{table*}

\begin{table}[t]	
	\centering\renewcommand\arraystretch{1.2}\setlength{\tabcolsep}{8pt}
	\belowrulesep=0pt\aboverulesep=0pt
	\caption{Cross-manipulation evaluation results of methods trained only on real face videos from FF++. \label{cross-mani}}
    \vspace{-3mm}
	\begin{tabular}{c|cccc|c}
		\toprule
		\multirow{2}{*}{Method} & 
		\multicolumn{5}{c}{Test Set AUC (\%)} \\
		\cmidrule(lr){2-6}
		&\multicolumn{1}{c}{DF} 
		&\multicolumn{1}{c}{F2F} 
		&\multicolumn{1}{c}{FS}
		&\multicolumn{1}{c|}{NT}
		&\multicolumn{1}{c}{FF++} \\
		\midrule
		SBI \cite{Shiohara_2022_CVPR} & \underline{99.99} & 99.88 & 99.91 & 98.79 & 99.64 \\
		SeeABLE \cite{Larue_2023_ICCV} & 99.20 & 98.80 & 99.10 & 96.90 & 98.50 \\
		AUNet \cite{Bai_2023_CVPR} & 99.98 & 99.60 & 99.89 & 98.38 & 99.46 \\
		RAE \cite{10.1007/978-3-031-72943-0_23} & 99.60 & 99.10 & 99.20 & 97.60 & 98.90 \\
        WMamba \cite{peng2025wmamba} & \textbf{100} & \underline{99.98} & \underline{99.94} & \underline{98.88} & \underline{99.70} \\
		\midrule
		SGF-CDNet & \textbf{100} & \textbf{100} & \textbf{99.96} & \textbf{98.91} & \textbf{99.74}\\      
		\bottomrule
	\end{tabular}
    \vspace{-4mm}
\end{table}

\subsubsection{Gated Fusion and Spatial Mapping}
To dynamically combine the information from these two complementary dimensions, we concatenate the consistency representation $H_\text{consis}$ and the discrepancy representation $H_\text{disc}$ along the feature dimension and then generate an adaptive gating weight $\lambda$ through an MLP and sigmoid function operation:
\begin{equation}
    \lambda = \sigma(\text{MLP}(\text{Concat}(H_\text{consis}, H_\text{disc}))). 
\end{equation}

The final graph reasoning result $H_\text{fused}$ is a weighted sum of the two path outputs. Subsequently, this fused result is added to the module's input node features $Z$ via a residual connection to obtain the final reasoned nodes $H_\text{reasoned}$. This process can be represented as:
\begin{equation}
    \begin{aligned}
    H_\text{fused} &= (1 - \lambda) \cdot H_\text{consis} + \lambda \cdot H_\text{disc},\\
    H_\text{reasoned} &= Z + \text{MLP}(H_\text{fused}).
    \end{aligned}
\end{equation}

Finally, we broadcast the reasoned node features $H_{\text{reasoned}} = \{h_1, \dots, h_N\}$ back to the original 2D space in the spatial mapper based on their corresponding one-hot encoded masks $\{M_{i,n}^\text{onehot}\}_{n=1}^{N}$ to form a structured gating factor map $G_\text{map}$:
\begin{equation}
    G_\text{map} = \sum_{n=1}^{N} h_n \cdot M_{i,n}^\text{onehot}.
\end{equation}

This map is finally applied to the image encoder's feature map, providing top-down structural guidance and enhancement for the underlying features.

\section{EXPERIMENTS}
\vspace{-2mm}
\subsection{Settings}
\vspace{-1mm}
\subsubsection{Datasets}

We use the FaceForensics++ (FF++) \cite{Rossler_2019_ICCV} dataset. Following the SBI \cite{Shiohara_2022_CVPR} training protocol, our model is trained on the 1000 authentic videos. For intra-dataset evaluation, the trained model is tested on videos generated by the four manipulation techniques in FF++: DeepFakes (DF), Face2Face (F2F), FaceSwap (FS) and NeuralTextures (NT). To assess the model's generalization capabilities, we perform cross-dataset evaluations on four public datasets. These include Celeb-DeepFake-v2 (CDF2) \cite{Li_2020_CVPR}, the DeepFake Detection Challenge (DFDC) \cite{dolhansky2020deepfake} and its Preview version (DFDCP) \cite{dolhansky2019dee} and FFIW-10K (FFIW) \cite{Zhou_2021_CVPR}.

\subsubsection{Baselines}
We compare our model's performance with 11 representative frame-level and video-level detection baselines, including SBI \cite{Shiohara_2022_CVPR}, SeeABLE \cite{Larue_2023_ICCV}, AUNet \cite{Bai_2023_CVPR}, LAA-Net \cite{Nguyen_2024_CVPR}, RAE \cite{10.1007/978-3-031-72943-0_23}, FreqBlender \cite{zhou2024freqblender}, UDD \cite{fu2025exploring}, WMamba \cite{peng2025wmamba}, TALL++ \cite{xu2024towards}, NACO \cite{zhang2025learning} and DFD-FCG \cite{11093395}. 

\subsubsection{Evaluation Metric}
To evaluate the detection performance, we use the Area Under the Receiver Operating Characteristic Curve (AUC). Since our method operates on a frame-level basis, we calculate the final score for each video by averaging the prediction scores from all of its individual frames. This protocol allows our model to be fairly and directly compared with both frame-level and video-level baselines.

\subsubsection{Implementation Details}
Our experimental setup directly follows the SBI framework \cite{Shiohara_2022_CVPR}, using its methods for synthetic data generation, preprocessing and data augmentation, detailed in the supplementary materials. For our model, we use a ConvNeXt-Base network \cite{9879745} as the image encoder, which has four stages with 3, 3, 27 and 3 blocks respectively. To start with, the encoder is loaded with weights pre-trained on the ImageNet-1K dataset. The number of facial semantic regions $N$ is 18, inheriting the annotation standards from the CelebAMask-HQ benchmark dataset after ignoring the ``background" category and we employ a pretrained BiSeNet \cite{yu2018bisenet} model to perform face parsing, which provides the semantic segmentation masks for different facial regions. The $\epsilon$ in (\ref{1}) is 1e-6. The number of landmarks $K$ per image is 68. The number of attention heads $H$ of the geometric attention module in SGFM and of the first GAT layers in consistency and discrepancy paths is 4. We train the model for 200 epochs with the AdamW optimizer and a batch size of 64. The learning rate starts at 5e-5 with a linear learning rate decay scheduled after the 100th epoch. The model is implemented using the PyTorch framework, consuming 20.7 GB of GPU memory during training.

\begin{table}[t]
    \centering\renewcommand\arraystretch{1.2}\setlength{\tabcolsep}{5.5pt}
	\belowrulesep=0pt\aboverulesep=0pt
	\caption{Ablation study results for different inputs in SGFM.  \label{table3}}
    \vspace{-3mm}
    \begin{tabular}{cc|cccc}
    \toprule
    \multicolumn{2}{c|}{Method} & \multicolumn{4}{c}{Test Set AUC (\%)} \\ \cmidrule(lr){1-2} \cmidrule(lr){3-6} 
    Parsing Results & Landmarks & CDF2 & DFDC & DFDCP & FFIW   \\ 
    \midrule
     \usym{2717} & \usym{2717} & 96.19 & 82.01 & 86.84 & \underline{86.36}\\
    \usym{2713}  & \usym{2717} & \underline{96.37} & \underline{84.21} & \underline{89.72} & 82.75 \\
 \midrule
 \usym{2713} & \usym{2713} &  \textbf{97.37} & \textbf{84.72} & \textbf{89.77}   & \textbf{87.23}   \\ \bottomrule
\end{tabular}
\vspace{-6mm}
\end{table}

\begin{table}[t]	
    \centering\renewcommand\arraystretch{1.2}\setlength{\tabcolsep}{4pt}
	\belowrulesep=0pt\aboverulesep=0pt
	\caption{Ablation study results for CD-GNN. ``Consistency'' means the consistency path and ``Discrepancy'' means the discrepancy path.\label{table4}}
    \vspace{-3mm}
    \begin{tabular}{ccc|cccc}
    \toprule
    \multicolumn{3}{c|}{Method} & \multicolumn{4}{c}{Test Set AUC (\%)} \\ \cmidrule(lr){1-3} \cmidrule(lr){4-7} 
    GNN & Consistency &Discrepancy & CDF2 & DFDC & DFDCP & FFIW   \\ 
    \midrule
    \usym{2717} & \usym{2717} & \usym{2717} & 95.95 & 82.44 & 87.89 & 85.24\\
    \usym{2713} & \usym{2717} & \usym{2717} & 96.04 & 82.94 & \underline{89.73} & 85.83\\
   \usym{2713} & \usym{2713}  & \usym{2717} & \underline{96.55} & 83.72 & 88.04 & 82.99 \\
   \usym{2713} & \usym{2717} & \usym{2713} & 94.92 & \underline{84.01} & 88.91 & \underline{86.76}
      \\ \midrule
\usym{2713} & \usym{2713} & \usym{2713} &  \textbf{97.37} & \textbf{84.72} & \textbf{89.77}   & \textbf{87.23}   \\ \bottomrule
\end{tabular}
\vspace{-4mm}
\end{table}

\begin{table}[t]
\centering\renewcommand\arraystretch{1.2}\setlength{\tabcolsep}{10.2pt}
	\belowrulesep=0pt\aboverulesep=0pt
	\caption{Ablation study results for fusion strategy between the outputs of CD-GNNs and features from the image encoder. \label{table5}}
    \vspace{-3mm}
	\begin{tabular}{c|cccc}
		\toprule
		\multirow{2}{*}{Method} & 
		\multicolumn{4}{c}{Test Set AUC (\%)} \\
		\cmidrule(lr){2-5}
		&\multicolumn{1}{c}{CDF2} 
		&\multicolumn{1}{c}{DFDC} 
		&\multicolumn{1}{c}{DFDCP}
		&\multicolumn{1}{c}{FFIW} \\
		\midrule
		 Addition  & \underline{94.82} & \underline{82.56} & \underline{88.06} & \underline{84.58} \\
		 Product  & 92.54 & 80.50 & 85.73 & 73.24 \\
		 Concatenation  & 91.49 & 78.61 & 80.49 & 70.23 \\
		\midrule
		Gating & \textbf{97.37} & \textbf{84.72} & \textbf{89.77} & \textbf{87.23}\\ 
		\bottomrule
	\end{tabular}
    \vspace{-7mm}
\end{table}

\vspace{-2mm}
\subsection{Results}
\vspace{-1mm}
\subsubsection{Cross-Dataset Evaluation}
To assess our model's generalization ability, we perform a cross-dataset evaluation. The detailed results of this and the performance of recent SOTA methods are presented in Table \ref{cross-data}. The proposed SGF-CDNet achieves the highest AUC scores across all unseen datasets, proving its superior generalization capability. This promising performance can be attributed to the high quality node presentations in SGFM and the effectiveness of feature structure analysis in CD-GNN, which fully leverage structural information of relationships between facial components.

\subsubsection{Cross-Manipulation Evaluation}
In practice, the specific technique used to generate deepfake images is rarely known in advance. This makes it essential for a detector to perform well against manipulation types it has not seen during training. To evaluate this, we follow the protocol from \cite{Shiohara_2022_CVPR}, where the model is trained only on real faces from FF++ and then tested on different forgery methods. As shown in Table \ref{cross-mani}, our SGF-CDNet is highly effective in this setting.

\vspace{-2mm}
\subsection{Ablation Studies}
\vspace{-1mm}
\subsubsection{Analysis of SGFM} We conduct ablation studies on the information inputs of SGFM, as shown in Table \ref{table3}. Since it is impossible to generate the necessary nodes for the CD-GNN without using both face parsing results and landmarks, the performance for this configuration is reported using the baseline ConvNeXt model alone, without any of the modules proposed in this paper. The results show that the performance improve significantly after using face parsing. With the addition of landmarks, the performance is further enhanced. This demonstrates the effectiveness of both the face parsing results in guiding the semantic nodes and the geometric information provided by the landmarks.

\begin{figure}[thpb]
  \centering
\includegraphics[width=6cm]{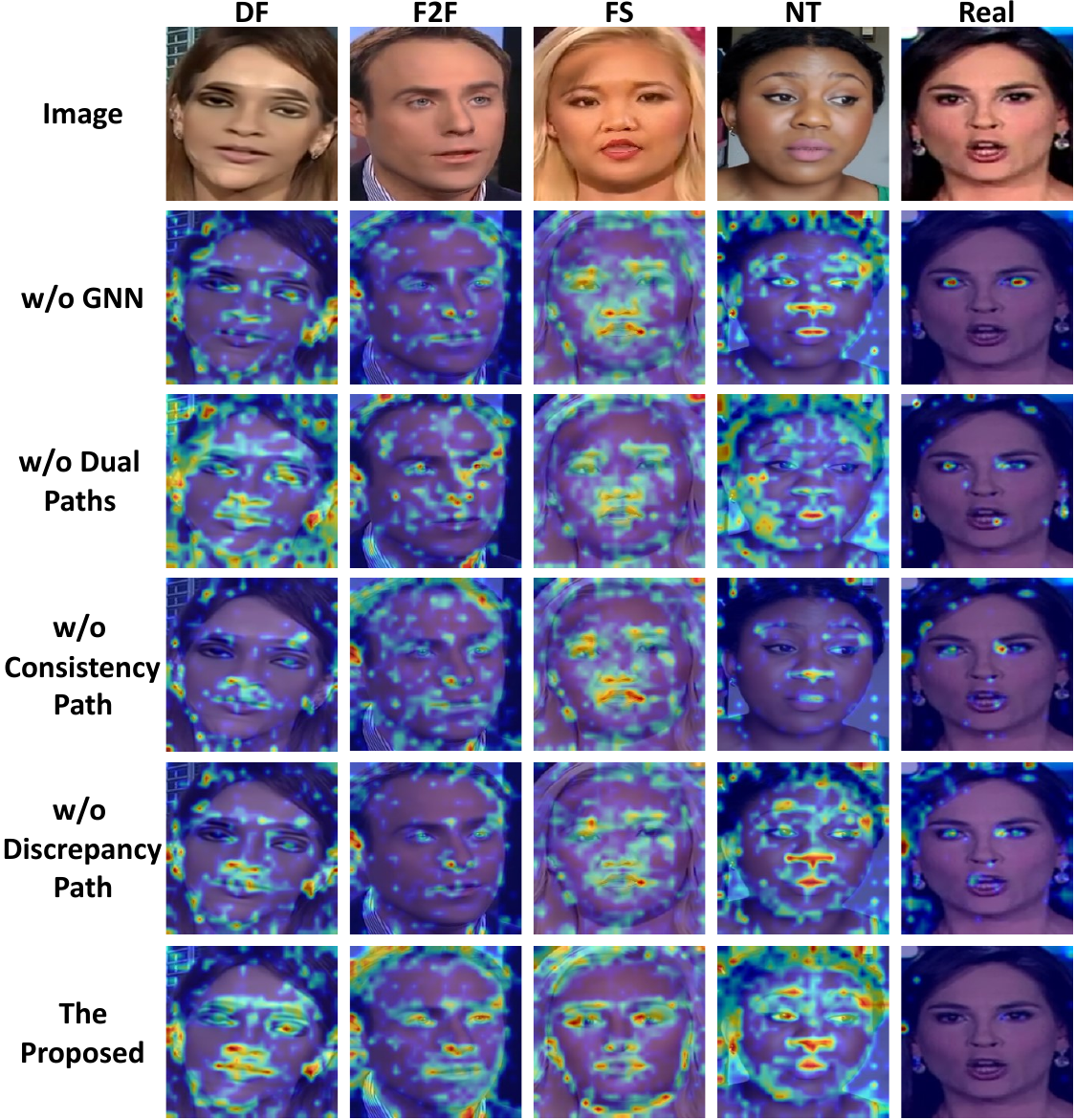}
\vspace{-3mm}
  \caption{The saliency maps are shown for one real sample and four distinct manipulated samples.}
  \vspace{-7mm}
  \label{figure4}

\end{figure}

\subsubsection{Analysis of CD-GNN} Table \ref{table4} presents ablation studies on the CD-GNN architecture. Without graph reasoning, semantic and geometric features are underutilized, leaving poor cross-dataset performance. While partial configurations (single-path or GNN-only) achieve sub-optimal results, our full model optimally fuses nodes from both consistency and discrepancy paths, achieving the best overall performance and validating the dual-path design.


\subsubsection{Effectiveness of Gating Mechanism} To validate the effectiveness of gating mechanism, we compare its performance against  element-wise addition, product and channel-wise concatenation, as shown in Table \ref{table5}. The cross-dataset evaluation results, show the superiority of gating mechanism.

\subsubsection{Visual Analysis of SGF-CDNet} Grad-CAM visualizations (Fig. \ref{figure4}) further validate SGF-CDNet. Guided by semantic and geometric information, the model accurately focuses on key facial structures to highlight disharmony in manipulated images. Conversely, real images exhibit low, diffuse activations, confirming the framework's overall effectiveness. More visualization results are in supplementary materials.

\section{CONCLUSION}
We introduce SGF-CDNet, a structured graph reasoning framework that uncovers structural dependencies in facial features. By deeply fusing facial semantic regions with geometric landmarks, we significantly enhance node expressiveness. Our core contribution, the dual-path Consistency-Discrepancy Graph Network (CD-GNN), performs relational reasoning to capture facial disharmony effectively. Extensive experiments across multiple datasets demonstrate its superior performance and generalization ability.

\bibliographystyle{IEEEbib}
\bibliography{icme2026references}


\end{document}


\title{SGF-CDNet: A Consistency-Discrepancy Graph Network over Semantic-Geometric Fused Nodes for Face Forgery Detection \\
(Supplementary Material)}


\author{
\parbox{16cm}{\centering
    {\large Jiayao Jiang\textsuperscript{1,2}, Bin Liu\textsuperscript{1,2\dag}, Nenghai Yu\textsuperscript{1,2}}\\
    {\normalsize
    \textsuperscript{1}School of Cyber Science and Technology, University of Science and Technology of China, China\\
    \textsuperscript{2}Anhui Province Key Laboratory of Digital Security, China\\
    \thanks{\dag Corresponding author: \href{mailto:flowice@ustc.edu.cn}{flowice@ustc.edu.cn}}}
}}

\maketitle


\begin{abstract}
In this material, we detail the implementation specifics of our method. Then, we introduce the related work of our paper. Finally, we provide the extended experimental results to further prove the effectiveness of our design.
\end{abstract}


\section{Implementation Details}
In this work, we follow the methodology of SBI \cite{Shiohara_2022_CVPR}, adopting its synthetic data generation framework, preprocessing techniques and data augmentation strategies.

\subsection{Synthetic Data Generation}
The SBI framework generates synthetic training data by emulating the pipeline of deepfake generation. This process is driven by two core modules operating in parallel: the Source-Target Generator (STG) and the Mask Generator (MG). First, a real face image is fed into the STG, which applies a sequence of image transformations to create a pair of images: a pseudo-source and a pseudo-target. Simultaneously, the same input image is processed by the MG to produce a blending mask. This mask is constructed based on pre-detected facial landmarks and subjected to augmentation to enhance the diversity of the output. In the final step, these components are integrated. The pseudo-source and pseudo-target images are combined using the augmented blending mask, resulting in the final pseudo-fake face sample.

\subsection{Preprocessing}
Our data preprocessing approach varies between the training and inference stages to optimize performance. In the training phase, there are two key processes. First, we utilize RetinaFace \cite{Deng_2020_CVPR} to generate facial bounding boxes. Based on these boxes, each face is cropped with a random margin varying from 4\% to 20\% to augment the data. Second, Dlib's 81-point shape predictor \cite{10.5555/1577069.1755843} is used to extract facial landmarks, a step that is performed exclusively during training. For the inference phase, the process is simplified. While RetinaFace is still used to obtain the facial bounding boxes, a deterministic fixed margin of 12.5\% is applied for cropping. The extraction of facial landmarks is omitted during inference as it is not required for the final prediction.

\subsection{Data Augmentation}
All data augmentations are implemented using the versatile image processing toolbox from Albumentations \cite{buslaev2020albumentations}, applied at two distinct stages of our methodology. The first application occurs within the Source-Target Generator (STG), where transformations such as RGBShift, HueSaturationValue, RandomBrightnessContrast, Downscale and Sharpen are utilized to produce the pseudo-source and pseudo-target images. Subsequently, during the main training phase, the entire dataset, including both real and synthetic samples, undergoes another set of augmentations. This training-time process involves ImageCompression along with further color-space adjustments like RGBShift, HueSaturationValue and RandomBrightnessContrast to increase data diversity and improve model robustness.

\subsection{Other Details}
During the training phase, we sample eight frames from each video. In the testing phase, this number is increased to 32 frames per video. All input images are uniformly resized to a resolution of $224\times224$ for ConvNeXt-Base \cite{9879745} and Swin Transformer-Base \cite{Liu_2021_ICCV} image encoders during both training and testing. For EfficientNet-B4 \cite{tan2019efficientnet} image encoder, the size of input images is resized to a resolution of $380\times380$.  For frames that contain multiple detected faces, the classifier assesses each face independently, and the highest forgery score is selected as the predicted confidence for that frame.

\section{Related Work and Motivation}

\subsection{Face Forgery Detection}

Face forgery detection is fundamentally a binary classification task, aiming to determine whether a given image or video is authentic or manipulated. The research in this area can be effectively organized into four categories, as surveyed in \cite{pei2024deepfake}: spatial-domain, time-domain, frequency-domain and data-driven methods. Spatial-domain approaches directly scrutinize image-level evidence, such as artifacts \cite{Zhao_2021_CVPR,Cao_2022_CVPR,Shiohara_2022_CVPR,9694644}, color \cite{8803740} and saturation \cite{8803661}. In contrast, time-domain methods leverage video sequences to identify temporal discontinuities and unnatural motion between frames \cite{Haliassos_2021_CVPR,10.1145/3474085.3475508,gu2022delving,gu2022hierarchical,10054130,10478974,zhang2025learning}. The third category operates in the frequency-domain, employing transformations like Fourier Transform (FFT) \cite{tan2024frequency}, Discrete Cosine Transform (DCT) \cite{Li_2021_CVPR,qian2020thinking} and Discrete Wavelet Transform (DWT) \cite{9447758,Liu_2022_ACCV,10.1145/3503161.3547832,10004978} to reveal subtle manipulation traces that are often subtle in the spatial domain. Finally, data-driven strategies focus on enhancing the learning process itself through advanced model architectures and optimized training paradigms \cite{Huang_2023_CVPR,Guo_2023_CVPR,Zhai_2023_ICCV,Yan_2023_ICCV,Yan_2024_CVPR}.

\subsection{Graph Neural Networks}

Graph Neural Networks (GNNs) \cite{4700287} represent a class of deep learning models specifically designed to operate on graph-structured data, where entities and their relationships are explicitly modeled as nodes and edges. A major advancement came with the development of the Graph Convolutional Network (GCN) \cite{DBLP:conf/iclr/KipfW17}, which generalized the highly successful convolution operation from regular grids like images to irregular graph structures. To further enhance the model's expressiveness, the Graph Attention Network (GAT) \cite{DBLP:conf/iclr/VelickovicCCRLB18} incorporated an attention mechanism, enabling nodes to dynamically weigh the importance of their neighbors during information aggregation. By constructing a graph of the face, the combination of face forgery detection and GNNs allows for better capture of the structural and temporal inconsistencies introduced by forgeries. This method can model deep relationships between facial regions, improving the generalizability and explainability of the detection. FADE \cite{tan2023deepfake} formalizes forgery detection as a graph classification task, employing Graph Convolutional Networks (GCNs) to discover forgeries by modeling biological consistency among facial action units. SFDG \cite{10203558} introduced a dynamic graph learning approach that models the relationships between image patches in both spatial and frequency domains to adaptively uncover forgery clues and improve generalization.

\subsection{Motivation}
A real face has a natural and complex structure. Even when a fake face has no obvious artifacts, it remains difficult to replicate the authentic and natural features of the underlying structure on real human faces. As tools designed specifically for relational reasoning, Graph Neural Networks (GNNs) are excellent choices for achieving a comprehensive analysis of the intrinsic structural relationships of a face. However, the power of a GNN depends on the quality of its node representations. Using only sparse facial landmarks as nodes leads to the loss of regional semantics and texture information. Conversely, using only semantic regions derived from face parsing lacks precise geometric constraints. Therefore, it is reasonable to deeply integrate these two sources of information, which come from semantic regions from face parsing results and augmented with geometric information from landmarks, to form the SGF nodes. The SGF nodes provide a high information density for the subsequent graph reasoning. Moreover, structural errors introduced by forgeries are twofold: they violate the expected consistency among facial components and introduce anomalous discrepancies. A single reasoning path can hardly capture both of these complementary signals simultaneously. Therefore, we design a dual-path CD-GNN framework with consistency and discrepancy paths. The consistency path is responsible for verifying whether facial components conform to natural harmonious patterns, while the discrepancy path actively seeks out the unnatural conflicts introduced by forgeries. This dual-path reasoning is the key to understanding the structural integrity of a face deeply. Finally, we compose SGF nodes and CD-GNN to derive the SGF-CDNet framework, which is an effective and reliable solution for face forgery detection.

\begin{figure*}[thpb]
  \centering 
  \includegraphics[width=\textwidth]{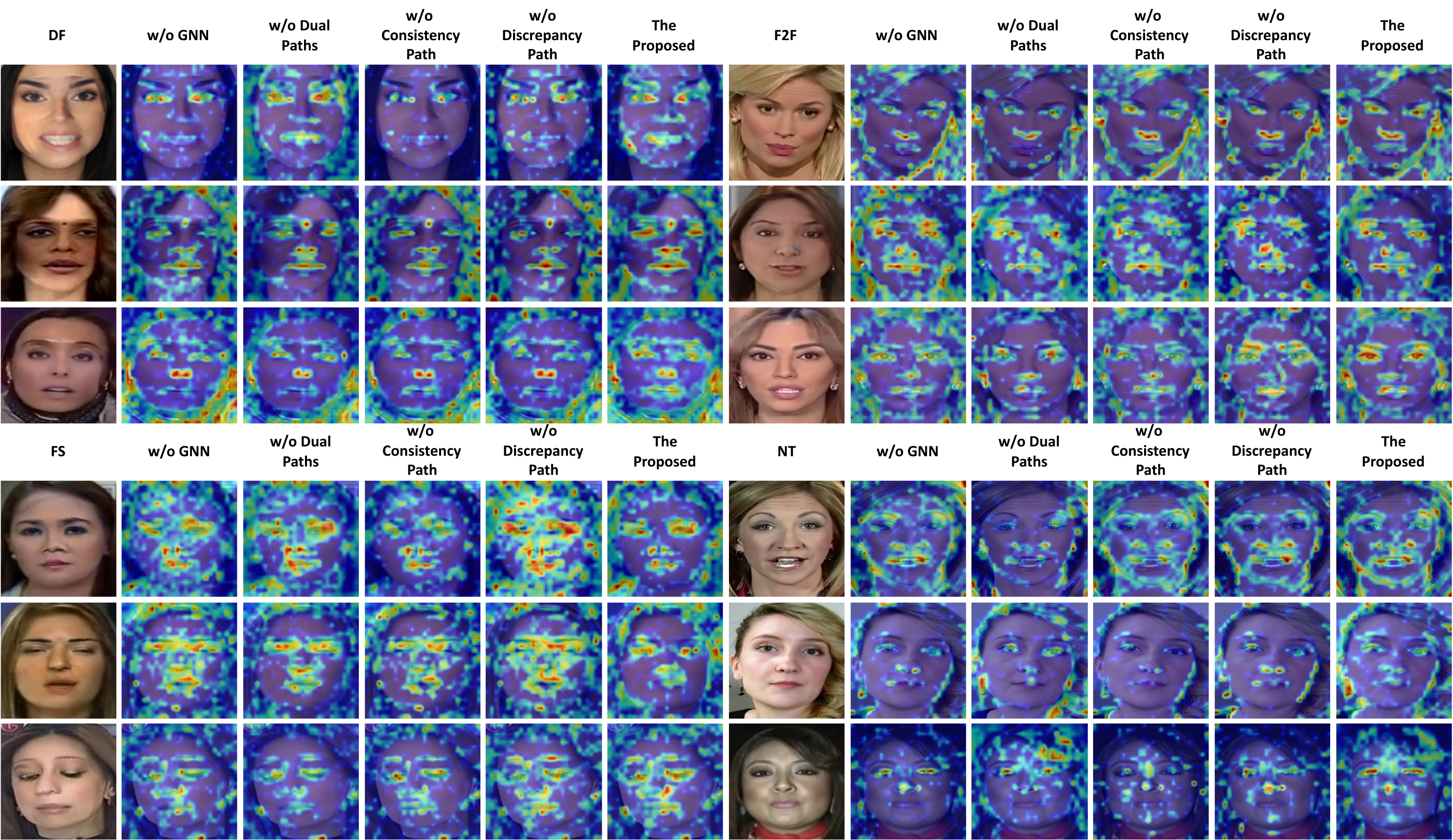}
  \caption{Saliency maps are visualized for a selection of fake samples from the FF++ dataset. The intensity of the red overlay corresponds to the degree of model attention. The CD-GNN enhances the model's ability to focus on the harmonious and disharmonious relationships between facial components.}
  \label{figure1}
\end{figure*}

\section{Experimental Results}
In this section, we provide supplementary experimental results that are excluded from the main manuscript.

\begin{table}[t]
    \centering\renewcommand\arraystretch{1.2}\setlength{\tabcolsep}{6pt}
	\belowrulesep=0pt\aboverulesep=0pt
	\caption{Ablation study results for choices on using the SGF-CDNet at different stages of the image encoder. ``Avg.'' is short for average. \label{table21}}
    \begin{tabular}{cccc|cccc|c}
    \toprule
    \multicolumn{4}{c|}{Injection Stage} & \multicolumn{5}{c}{Test Set AUC (\%)} \\ \cmidrule(lr){1-4} \cmidrule(lr){5-9} 
    1 & 2 & 3 & 4 & CDF2 & DFDC & DFDCP & FFIW  & Avg. \\ 
    \midrule
   \usym{2717} & \usym{2717} & \usym{2717} & \usym{2717} & 96.19 & 82.01 & 86.84 & 86.36 & 87.85 \\
   \usym{2713} & \usym{2717} & \usym{2717} & \usym{2717} & 94.89 & 84.38 & 89.68 & 85.31 & 88.57\\
  \usym{2713} & \usym{2713} & \usym{2717}  & \usym{2717} & 95.49 & 82.58 & 89.24 & 82.72 & 87.51\\
  \usym{2713} & \usym{2713} & \usym{2713} & \usym{2717} & 95.38 & 83.68 & \textbf{89.92} & 83.51 & 88.12
      \\ \midrule
\usym{2713} & \usym{2713} & \usym{2713} & \usym{2713} &  \textbf{97.37} & \textbf{84.72} & 89.77   & \textbf{87.23}  & \textbf{89.77} \\ \bottomrule
\end{tabular}
\end{table}

\begin{table}[t]	
	\centering\renewcommand\arraystretch{1.2}\setlength{\tabcolsep}{4.3pt}
	\belowrulesep=0pt\aboverulesep=0pt
	\caption{Performance and efficiency analysis of popular network architectures with and without SGF-CDNet. ``Params.'' is short for parameters and ``Avg.'' is short for average. \label{table22}}
	\begin{tabular}{c|c|cccc|c}
		\toprule
		\multirow{2}{*}{Method} & 
		\multirow{2}{*}{Params.} &  
		\multicolumn{5}{c}{Test Set AUC (\%)} \\
		\cmidrule(lr){3-7}
		&\multicolumn{1}{c|}{} 
		&\multicolumn{1}{c}{CDF2} 
		&\multicolumn{1}{c}{DFDC} 
		&\multicolumn{1}{c}{DFDCP}
		&\multicolumn{1}{c|}{FFIW}
        &\multicolumn{1}{c}{Avg.}\\
		\midrule
		EfficientNet-B4 & 17.6M  & 93.18 & 72.42 & 86.15 & 84.83 & 84.15 \\
		+ SGF-CDNet & 18.9M  & 94.83 & 85.18 & 87.36 & 86.26 & 88.41 \\
		\midrule
		Swin-B & 86.7M & 95.53 & 82.81 & 89.70 & 85.38 & 88.36 \\
		+ SGF-CDNet & 104.1M & 95.63 & \textbf{85.58} & \textbf{91.06} & 84.38 & 89.16 \\
		\midrule
		ConvNeXt-B & 87.6M & 96.19 & 82.01 & 86.84 & 86.36 & 87.85\\ 
		+ SGF-CDNet & 105.0M & \textbf{97.37} & 84.72 & 89.77 & \textbf{87.23} & \textbf{89.77}\\ 
		\bottomrule
	\end{tabular}
\end{table}

\subsection{Analysis on the Injection Stage of SGF-CDNet} We also conducted experiments on injecting SGF-CDNet at different stages of the image encoder, as shown in Table \ref{table21}. The results show that although the baseline with no injection performs well on some datasets, injecting SGF-CDNet at all stages achieves comparatively superior results.

\subsection{Analysis of Different Image Encoders} As shown in Table \ref{table22}, we conduct experiments on several different image encoders, including EfficientNet \cite{tan2019efficientnet} and Swin Transformer \cite{Liu_2021_ICCV}. ``Param.'' denotes the number of trainable parameters. Notably, the weights of the BiSeNet, which generates the face parsing results, are frozen and not included in this count. For each image encoder, the SGF-CDNet module is inserted at every stage after each feature downsampling layer, which is similar to the ConvNeXt-Base image encoder. The experimental results show that SGF-CDNet provides a significant boost in generalization performance for all tested image encoders, while increasing the original model's parameter count by less than 20\%. Although other image encoders surpass ConvNeXt's generalization performance on some individual datasets, selecting ConvNeXt as the image encoder achieves the best average performance across all datasets. Therefore, ConvNeXt is chosen as the image encoder for the main experiments in this paper. This experiment demonstrates that our SGF-CDNet is a lightweight, plug-and-play module that effectively enhances the performance for a variety of different image encoders.

\begin{figure}[t]
  \centering
\includegraphics[width=8.5cm]{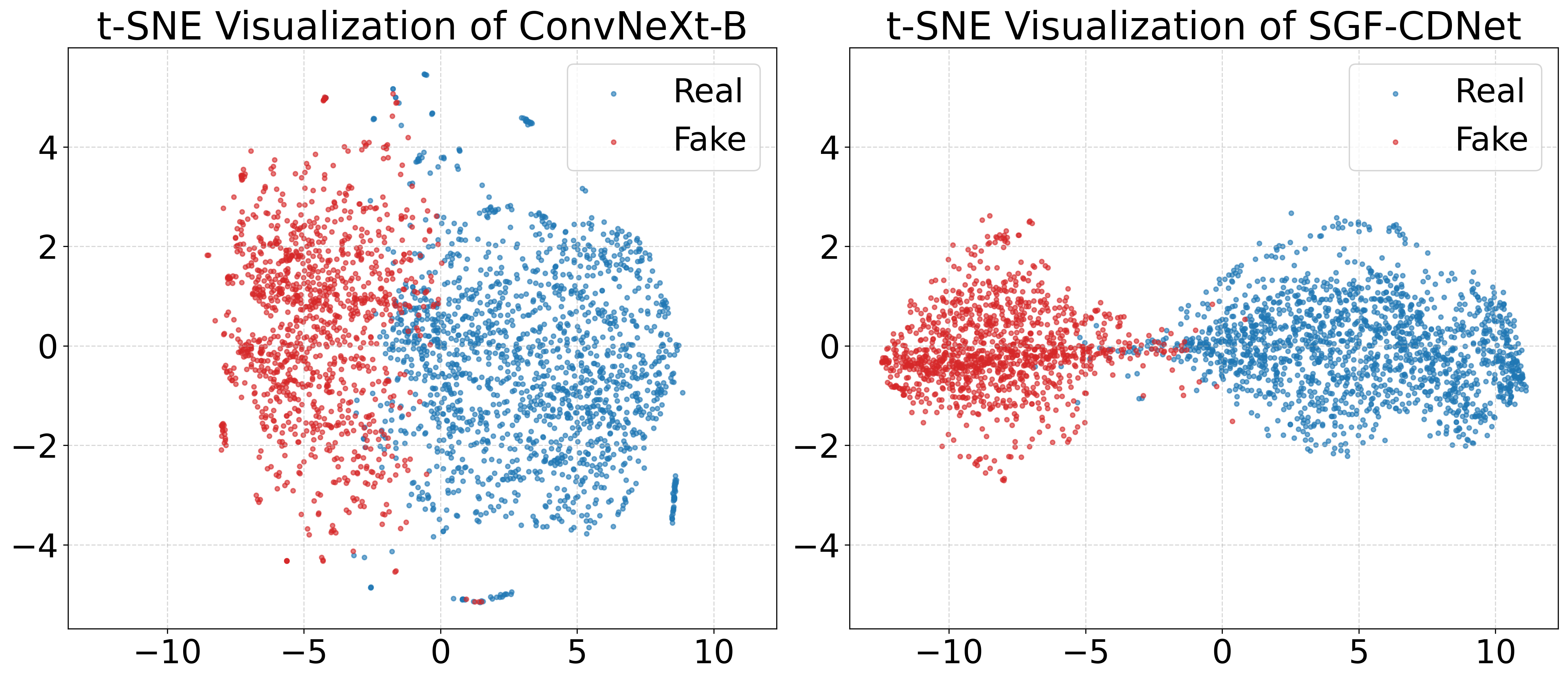}
\vspace{-2.5mm}
  \caption{A comparison of the t-SNE feature visualizations from the CDF2 dataset reveals the superior discriminative power of the proposed SGF-CDNet. In contrast to the baseline ConvNeXt, our model's features form clusters that are significantly more defined and clearly separated.}
  \label{figure23}
\vspace{-4mm}
\end{figure}

\subsection{Additional Analysis of CD-GNN and SGF-CDNet}
The CD-GNN analyzes the relationships between SGF nodes from complementary perspectives of consistency and discrepancy through two parallel paths. As visually demonstrated in Fig. 4 in the main text, using Graph Neural Networks (GNN) with consistency and discrepancy paths is crucial in directing model's attention toward key facial structures. To further validate this observation, we present additional visualization results. As illustrated in Fig. \ref{figure1}, the dual-path approach enhances the model's ability to focus on the intrinsic structural relationships during decision-making. Moreover, t-SNE visualizations in Fig. \ref{figure23} further prove that SGF-CDNet promotes a more distinct separation between real and fake samples than using the base detector alone, showing its higher generalization performance.

\subsection{Model Robustness}

\begin{table}[t]
	\centering
	\renewcommand\arraystretch{1.2}\setlength{\tabcolsep}{9pt}
	\belowrulesep=0pt\aboverulesep=0pt
	\caption{Comparison of robustness between SGF-CDNet and SBI \cite{Shiohara_2022_CVPR} models under different types and severities of image degradation.\label{degrad}}
	\begin{tabular}{c|c|ccc}
		\toprule
		\multirow{2}{*}{{Degradation}} & \multirow{2}{*}{{Method}} & \multicolumn{3}{c}{$\Delta$AUC by Severity (\%)} \\
		\cmidrule(lr){3-5}
		& & {Low} & {Medium} & {High} \\
		\midrule
		\multirow{2}{*}{{Compression}} 
		& {SBI \cite{Shiohara_2022_CVPR}} & -0.73 & -1.66 & -2.29 \\
		& {SGF-CDNet} & \textbf{-0.61} & \textbf{-0.94} & \textbf{-1.27} \\
		\midrule
		\multirow{2}{*}{{Occlusion}} 
		& {SBI \cite{Shiohara_2022_CVPR}} & -24.71 & -32.31 & -36.08 \\
		& {SGF-CDNet} & \textbf{-18.76} & \textbf{-28.10} & \textbf{-33.35} \\
		\midrule
		\multirow{2}{*}{{Blur}} 
		& {SBI \cite{Shiohara_2022_CVPR}} & -3.20 & -12.77 & -20.30 \\
		& {SGF-CDNet} & \textbf{-2.87} & \textbf{-12.69} & \textbf{-20.25} \\
		\midrule
		\multirow{2}{*}{{Noise}} 
		& {SBI \cite{Shiohara_2022_CVPR}} & -14.57 & \textbf{-29.75} & \textbf{-31.33} \\
		& {SGF-CDNet} & \textbf{-11.90} & {-34.01} & {-36.09} \\
		\bottomrule
	\end{tabular}
\end{table}

\subsubsection{Robustness for Real-world Degradation Scenarios}To verify the robustness of our approach, we perform evaluations under four distinct real-world degradation scenarios: compression, occlusion, blur and noise. We simulated compression using the JPEG standard at low (quality 95), medium (quality 85) and high (quality 75) settings. For occlusion, we introduce degradation by randomly masking 2\% (low), 5\% (medium) and 10\% (high) of total image pixels. Blur is applied via a Gaussian filter with kernel/sigma values set to 5/1 for low, 7/2 for medium and 9/3 for high levels of distortion. For noise, we introduce Gaussian noise with standard deviations of 0.01 (low), 0.05 (medium) and 0.1 (high). The cross-dataset results on the CDF2\cite{Li_2020_CVPR} dataset, detailed in Table~\ref{degrad}, show that SGF-CDNet consistently surpasses SBI with a significant margin in compression, occlusion and blur degradation scenarios. In the noise scenario, our model exhibits lower performance robustness than SBI under moderate and high noise conditions. This is likely because the degradation severely affects the representation quality of the graph nodes, which are fundamental to our structural graph reasoning. 

\begin{table}[t]	
\centering\renewcommand\arraystretch{1.2}\setlength{\tabcolsep}{9.5pt}
	\belowrulesep=0pt\aboverulesep=0pt
	\caption{Performance of SGF-CDNet and baseline models under different yaw angles for head poses on CDF2 dataset.\label{table7}}
	\begin{tabular}{c|c|c}
		\toprule
		\multicolumn{1}{c|}{Yaw Range} & 
		\multicolumn{1}{c|}{ConvNeXt-B AUC (\%)} &  
		\multicolumn{1}{c}{SGF-CDNet AUC (\%)} \\
        \midrule
		$0^\circ - 10^\circ$ & 95.99  &  96.59 \\
		$10^\circ - 20^\circ$ & 96.53 &  98.88\\
		$20^\circ - 30^\circ$ & 100 & 100 \\ 
        $> 30^\circ$  & N/A & N/A \\ 
		\bottomrule
	\end{tabular}
\end{table}

\begin{table}[t]	
\centering\renewcommand\arraystretch{1.2}\setlength{\tabcolsep}{9.5pt}
	\belowrulesep=0pt\aboverulesep=0pt
	\caption{Performance of SGF-CDNet and baseline models under different pitch angles for head poses on CDF2 dataset.\label{table8}}
	\begin{tabular}{c|c|c}
		\toprule
		\multicolumn{1}{c|}{Pitch Range} & 
		\multicolumn{1}{c|}{ConvNeXt-B AUC (\%)} &  
		\multicolumn{1}{c}{SGF-CDNet AUC (\%)} \\
        \midrule
		$0^\circ - 5^\circ$ & 93.93  &  97.55 \\
		$5^\circ - 10^\circ$ & 96.48 & 96.98 \\
		$10^\circ - 15^\circ$ & 97.02 & 97.54 \\ 
        $15^\circ - 20^\circ$ & 99.13 & 100 \\ 
        $> 20^\circ$  & N/A & N/A \\ 
		\bottomrule
	\end{tabular}
\end{table}

\subsubsection{Robustness for Head Pose Variations}To evaluate the robustness of SGF-CDNet under various head poses, we have conducted a fine-grained analysis on the CDF2 dataset across different pitch and yaw angle ranges. As summarized in Table~\ref{table7} and Table~\ref{table8}, our method consistently outperforms the ConvNeXt-Base baseline in nearly all pose intervals. Specifically, SGF-CDNet exhibits a significant advantage in the $0^\circ-5^\circ$ pitch range and the $10^\circ-20^\circ$ yaw range. Notably, our model maintains or even achieves a perfect AUC of 1.0 at certain larger angles, such as the $15^\circ-20^\circ$ pitch and $20^\circ-30^\circ$ yaw intervals. As for larger angle ranges, statistical analysis is not feasible due to the absence of corresponding facial samples in the CDF2 dataset. These results strongly demonstrate that our structural graph reasoning module, by effectively integrating semantic and geometric cues, captures invariant structural features that are resilient to viewpoint variations, whereas pure appearance-based models may struggle with texture distortions caused by rotation. 

\begin{table}[t]	
	\centering\renewcommand\arraystretch{1.2}\setlength{\tabcolsep}{9pt}
	\belowrulesep=0pt\aboverulesep=0pt
	\caption{Model performance under varying levels of Gaussian noise applied to landmark coordinates.\label{table2}}
	\begin{tabular}{c|c|c}
		\toprule
		\multicolumn{1}{c|}{Gaussian Noise Variance (pixel)} & 
		\multicolumn{1}{c|}{Test AUC (\%)} &  
		\multicolumn{1}{c}{$\Delta$AUC (\%)} \\
        \midrule
		$\sigma=0$ & 97.37  & 0.00  \\
		$\sigma=1$ & 97.37 & 0.00 \\
		$\sigma=2$ & 97.36 & -0.01 \\ 
        $\sigma=3$ & 97.36 & -0.01 \\ 
        $\sigma=5$ & 97.37 & 0.00 \\ 
        $\sigma=8$ & 97.37 & 0.00 \\ 
        $\sigma=10$ & 97.36 & -0.01 \\ 
		\bottomrule
	\end{tabular}
\end{table}

\begin{table}[t]	
	\centering\renewcommand\arraystretch{1.2}\setlength{\tabcolsep}{16pt}
	\belowrulesep=0pt\aboverulesep=0pt
	\caption{Model performance under varying levels of dropout applied to face parsing results.\label{table3}}
	\begin{tabular}{c|c|c}
		\toprule
		\multicolumn{1}{c|}{Parsing Drop Rate} & 
		\multicolumn{1}{c|}{Test AUC (\%)} &  
		\multicolumn{1}{c}{$\Delta$AUC (\%)} \\
        \midrule
		$p=0\%$ & 97.37  & 0.00  \\
		$p=5\%$ & 96.33 & -1.04 \\
		$p=10\%$ & 95.34 & -2.03 \\ 
        $p=20\%$ & 93.65 & -3.72 \\ 
        $p=30\%$ & 94.05 & -3.32 \\ 
        $p=50\%$ & 94.76 & -2.61 \\ 
		\bottomrule
	\end{tabular}
\end{table}

\subsubsection{Robustness for Pre-processing Perturbations}
To investigate the sensitivity of SGF-CDNet to the quality of auxiliary inputs, we conduct stress tests by introducing synthetic noise into the landmark coordinates and random dropouts into the face parsing masks. The results demonstrate that our model exhibits exceptional robustness to geometric and semantic perturbations.

Regarding geometric noise, we observe that the detection performance is remarkably stable, as shown in Table~\ref{table2}. When the landmark noise level, which is the Gaussian noise variance, increases from $\sigma=0$ to $\sigma=10$ pixels, the AUC on the CDF2 dataset only slightly fluctuates from 97.37\% to 97.36\%, remaining nearly constant even under significant coordinate shifts. This suggests that the attention-based fusion and graph reasoning modules can effectively rectify minor geometric misalignments.

As for semantic degradation, we evaluate the impact of the parsing drop rate $p \in [0\%, 50\%]$, as shown in Table~\ref{table3}. While increasing the dropout rate leads to a controlled reduction in performance, the model maintains a high AUC of 94.76\% even when 50\% of the semantic regions are randomly discarded. This performance decay indicates that CD-GNN does not over-rely on any specific facial region but instead aggregates robust structural cues from the entire facial graph. Such robustness is crucial for ensuring reliable performance in real-world scenarios where occlusions or low-quality images may compromise the accuracy of pre-processing tools.

\begin{figure}[t]
\centering  
\subfigure[]{
\includegraphics[width=0.45\textwidth]{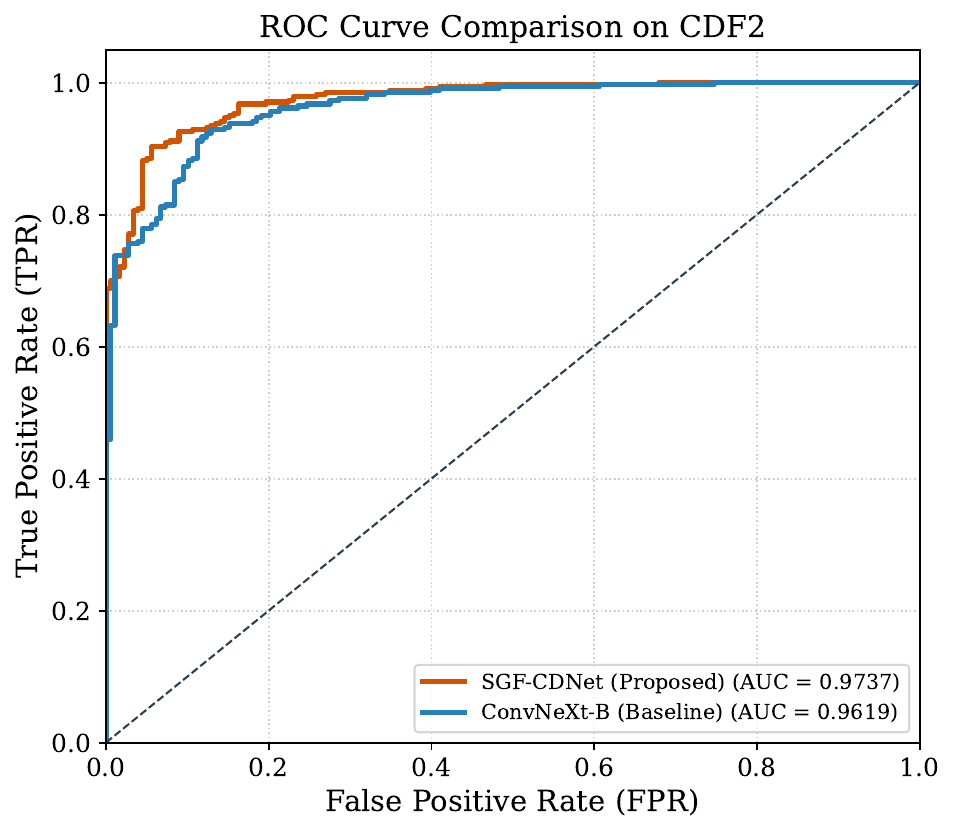}}
\subfigure[]{
\includegraphics[width=0.45\textwidth]{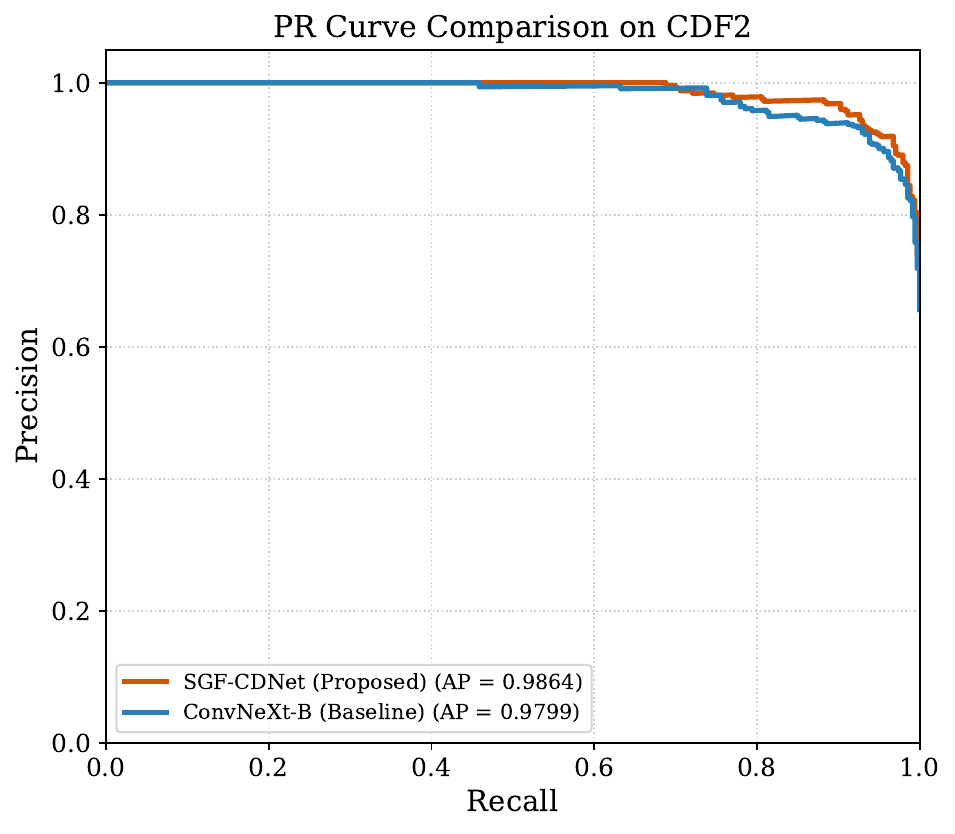}}
\caption{Graphical data: (a) ROC curves and (b) PR curves for SGF-CDNet and its backbone baseline on the CDF2 dataset.}
\label{figure2}
\end{figure}

\begin{figure}[t]
\centering  
\includegraphics[width=0.48\textwidth]{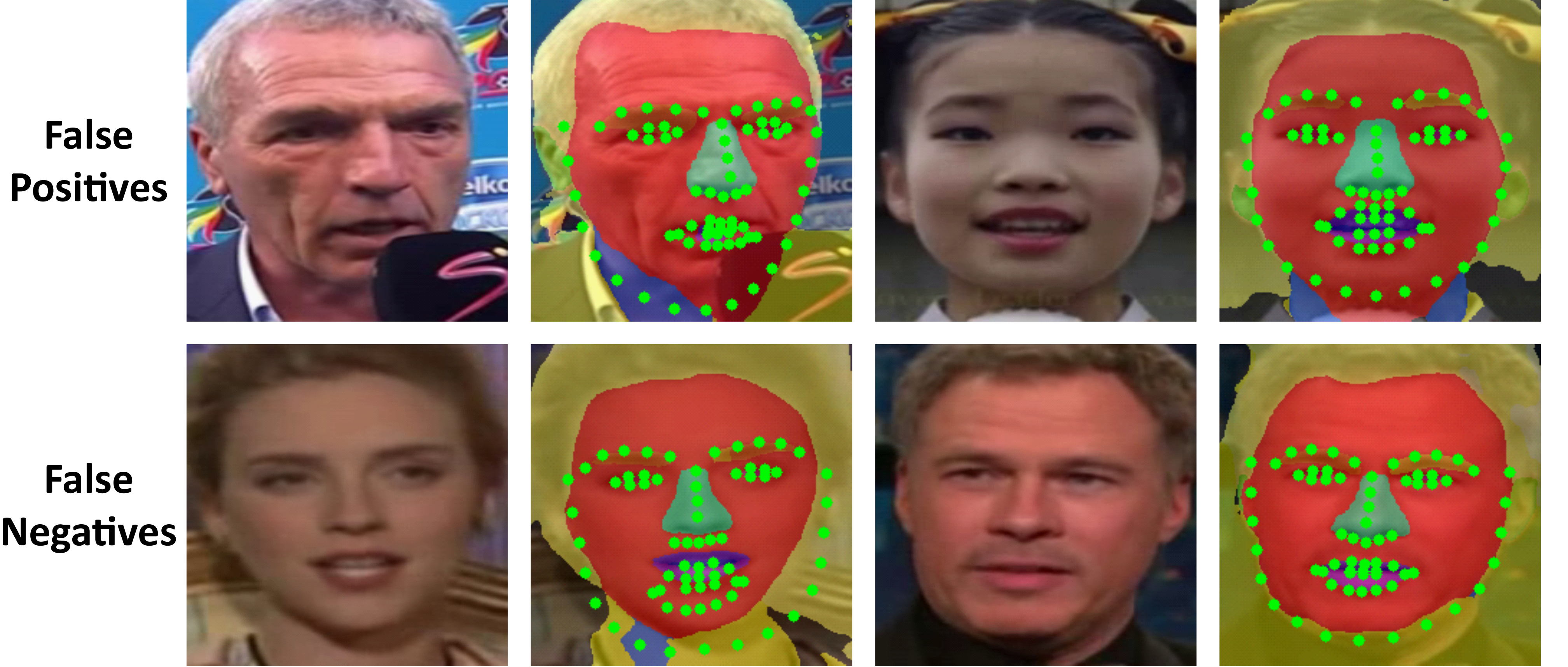}
\caption{Failure cases of frames and corresponding landmarks and face parsing results, including false positives and false negatives.}
\label{figure3}
\end{figure}

\begin{table}[t]
\centering\renewcommand\arraystretch{1.2}\setlength{\tabcolsep}{16.5pt}
	\belowrulesep=0pt\aboverulesep=0pt
	\caption{Performance of SGF-CDNet and baseline models under False Positive Rate (FPR) constraints. \label{table4}}
	\begin{tabular}{c|ccc}
		\toprule
		\multirow{2}{*}{Method} & 
		\multicolumn{3}{c}{True Positive Rate (\%)} \\
		\cmidrule(lr){2-4}
		&\multicolumn{1}{c}{0.1 FPR} 
		&\multicolumn{1}{c}{1 FPR} 
		&\multicolumn{1}{c}{10 FPR} \\
		\midrule
		 ConvNeXt-B  & 0.00 & 63.24 & 87.35 \\
		 +SGF-CDNet  & 0.00 & \textbf{70.00} & \textbf{92.65} \\
		\bottomrule
	\end{tabular}
\end{table}

\subsection{Performance Comparison and Failure Mode Analysis}
\subsubsection{Performance Comparison} To further evaluate the effectiveness of the proposed SGF-CDNet, we conduct a comprehensive comparison between the full SGF-CDNet and its backbone baseline (ConvNeXt-Base) on the CDF2 dataset. The Receiver Operating Characteristic (ROC) curve and Precision-Recall (PR) curve of SGF-CDNet and baseline are shown in Fig. \ref{figure2}. The SGF-CDNet's AUC score and average precision are 1.21\% and 0.65\% higher than the baseline, showing our proposed method further enhances the performance of the baseline model. Then, we compare the competitive performance of the models under strict False Positive Rate (FPR) constraints, as shown in Table~\ref{table4}. Our method shows the most significant improvement (+6.76\%) for True Positive Rate (TPR) at a strict FPR of 1.0\% and +5.3\% at an FPR of 10\%, demonstrating that our proposed model is capable of detecting more forged faces even under extremely low FPR constraints.

\subsubsection{Failure Mode Analysis} In our experiments, we observe a specific failure mode when evaluating under an extremely strict false positive rate (FPR=0.1\%). Both models report a TPR of 0.00\% with an infinite threshold. The occurrence of an infinite threshold at FPR=0.1\% indicates the presence of challenge cases within the dataset. In these cases, specific authentic frames are assigned abnormally high forgery scores due to extreme environmental factors, including occlusions, poor light conditions, etc. The original frames, landmarks and face parsing results for false positive outliers and false negative failures are illustrated in Fig. \ref{figure3}. The failure of the model in these instances is primarily attributed to the poor representation quality of graph nodes, which is caused by the misalignment of landmarks or inaccurate face parsing results.

\begin{table}[t]	
	\centering\renewcommand\arraystretch{1.2}\setlength{\tabcolsep}{3pt}
	\belowrulesep=0pt\aboverulesep=0pt
	\caption{Additional performance and efficiency analysis of popular network architectures with and without SGF-CDNet. Latency and inference speed results are measured on a single NVIDIA RTX 3090 GPU.\label{table5}}
	\begin{tabular}{c|c|c|c}
		\toprule
		\multicolumn{1}{c|}{Method} & 
		\multicolumn{1}{c|}{FLOPs} &  
		\multicolumn{1}{c|}{Average Latency (ms)} &
		\multicolumn{1}{c}{Inference Speed (FPS)} \\
		\midrule
		EfficientNet-B4 & 4.52G  & 20.24 & 49.40  \\
		+ SGF-CDNet & 18.36G  & 40.02 & 24.99  \\
		\midrule
		Swin-B & 15.17G & 15.41 & 64.90 \\
		+ SGF-CDNet & 29.14G & 52.69 & 18.98 \\
		\midrule
		ConvNeXt-B & 15.35G & 8.12 & 123.09\\ 
		+ SGF-CDNet & 29.25G & 38.00 & 26.32\\ 
		\bottomrule
	\end{tabular}
\vspace{1em}
	\centering
	\renewcommand\arraystretch{1.2} 
	\setlength{\tabcolsep}{7pt}      
	
	\caption{Efficiency analysis of SGF-CDNet with different backbones. ``Core FLOPs'' excludes the frozen BiSeNet. ``Loc.'' is short for localization. Preprocessing times are measured on an NVIDIA RTX 3090 GPU.}
	\label{table6}
	\belowrulesep=0pt
    \aboverulesep=0pt
	\begin{tabular}{c|c|c|c} 
		\toprule
		Backbone & 
		\begin{tabular}[c|]{@{}c@{}}Core FLOPs\end{tabular} & 
		\begin{tabular}[c|]{@{}c@{}}Landmark Loc. \\ Time (ms)\end{tabular} & 
		\begin{tabular}[c]{@{}c@{}}Face Parsing \\ Latency (ms)\end{tabular} \\
		\midrule
		EfficientNet-B4 & 4.53G & 6.10 & 5.36 \\
		Swin-B   & 15.31G & 4.85 & 6.59 \\
		ConvNeXt-B      & 15.43G & 4.89 & 5.67 \\
		\bottomrule
	\end{tabular}
\vspace{1em}
	\centering\renewcommand\arraystretch{1.2}\setlength{\tabcolsep}{9pt}
	\belowrulesep=0pt\aboverulesep=0pt
	\caption{Performance and efficiency comparison across different graph construction strategies on CDF2 dataset.\label{table9}}
	\begin{tabular}{c|c|c}
		\toprule
		\multicolumn{1}{c|}{Strategy} & 
		\multicolumn{1}{c|}{Training GPU Memory Cost} &  
		\multicolumn{1}{c}{Test AUC (\%)} \\
        \midrule
		Complete & 21.51 GB  & 95.42  \\
		Distance-driven & \textbf{19.84 GB} & 95.38 \\
		KNN & 22.50 GB & 95.70 \\ 
        Learned & 21.40 GB & 95.93 \\ 
        \midrule
        Dilation(Ours) & 20.74 GB & \textbf{97.37} \\ 
		\bottomrule
	\end{tabular}
\end{table}

\subsection{Additional Analysis of Computational Cost}

To provide a more comprehensive evaluation of SGF-CDNet's experimental performance, we have profiled the FLOPs, average latency and inference speed across various image encoders. As shown in Table~\ref{table5}, the integration of SGF-CDNet introduces additional computational overhead compared to the pure backbone architectures. For example, when employing ConvNeXt-Base as the backbone, the total FLOPs increase from 15.35G to 29.25G. However, despite the increased complexity, the model maintains a practical inference speed. Notably, the SGF-CDNet with ConvNeXt-Base achieves an inference speed of 26.32 FPS on a single NVIDIA RTX 3090 GPU, which is sufficient for real-time video-level forensic analysis.

Furthermore, we provide a detailed breakdown of the efficiency metrics in Table~\ref{table6} to highlight the source of the computational cost. We report the ``Core FLOPs'', which specifically excludes the frozen BiSeNet\cite{yu2018bisenet} preprocessing model. The number of frozen parameters of BiSeNet is 13.3M and its FLOPs is 13.84G. The results reveal that the structural graph reasoning module itself is highly efficient and lightweight, adding minimal burden to the overall pipeline. Additionally, we analyze the latency of preprocessing steps, including landmark localization and face parsing. It can be observed that these steps constitute a relatively constant overhead across different backbones. For instance, the face parsing latency consistently ranges from 5.36 ms to 6.59 ms, while landmark localization takes approximately 4.85 ms to 6.10 ms. These results demonstrate that the synergy between our structural reasoning module and existing backbones achieves a favorable trade-off between detection accuracy and computational efficiency.

\subsection{Evaluation of Alternative Graph Topologies} 
To justify our dilation-based design, we have conducted a comprehensive comparison on the CDF2 dataset against four alternatives: complete graph, distance-driven method, graph based on K-Nearest Neighbors (KNN)\cite{cover1967nearest} algorithm and graph based on learned dynamic adjacency. For the complete graph, it is a fully connected graph where every node pair possesses an edge. For the distance-driven method, edges are connected if the Euclidean distance between regional centroids is below a normalized threshold of $\tau=0.3$. For the graph based on KNN, each node is connected to its top $k=5$ most similar neighbors based on feature cosine similarity. For the graph based on learned dynamic adjacency, a lightweight MLP predicts edge weights between node pairs, activated by a threshold of $0.5$. The performance and efficiency comparison across different graph construction strategies on the CDF2 dataset are shown in Table~\ref{table9}. All training GPU memory costs reported are recorded with a batch size of 64. Our dilation-based approach achieves 97.37\% accuracy, significantly outperforming more flexible designs while maintaining a balanced GPU memory cost.

These results suggest that while complex or ``learned" connections seem more flexible, they often bring in irrelevant noise that hides the subtle signs of forgeries. In contrast, our approach follows real human facial structure by focusing only on neighboring facial parts. This ensures the model obtains exactly where structural errors are most likely to occur. This structural prior keeps the model focused on the most important areas, making it both easier to interpret and more accurate in detecting fake faces.

\bibliographystyle{IEEEbib}
\bibliography{icme2026references}
